\documentclass[11pt,a4paper]{article}
\usepackage[hyperref]{acl2021}
\usepackage{times}
\usepackage{latexsym}
\usepackage{amsmath}
\usepackage{graphicx}
\usepackage{multirow}
\usepackage{amssymb}
\usepackage{booktabs}

\usepackage{microtype}

\aclfinalcopy 

\title{Making Better Use of Bilingual Information for \\ Cross-Lingual AMR Parsing}

\author{Yitao Cai, Zhe Lin \and Xiaojun Wan \\
	Wangxuan Institute of Computer Technology, Peking University \\
	Center for Data Science, Peking University \\
	The MOE Key Laboratory of Computational Linguistics, Peking University \\
	{\tt \{caiyitao,linzhe,wanxiaojun\}@pku.edu.cn} \\}

\date{}

\begin{document}
\maketitle
\begin{abstract}
Abstract Meaning Representation (AMR) is a rooted, labeled, acyclic graph representing the semantics of natural language. As previous works show, although AMR is designed for English at first, it can also represent semantics in other languages. However, they find that concepts in their predicted AMR graphs are less specific. We argue that the misprediction of concepts is due to the high relevance between English tokens and AMR concepts. In this work, we introduce bilingual input, namely the translated texts as well as non-English texts, in order to enable the model to predict more accurate concepts. Besides, we also introduce an auxiliary task, requiring the decoder to predict the English sequences at the same time. The auxiliary task can help the decoder understand what exactly the corresponding English tokens are. Our proposed cross-lingual AMR parser surpasses previous state-of-the-art parser by 10.6 points on Smatch F1 score. The ablation study also demonstrates the efficacy of our proposed modules. 
\end{abstract}

\section{Introduction}

Abstract Meaning Representation (AMR) \cite{banarescu2013abstract} is a rooted, labeled, acyclic graph representing sentence-level semantic of text. Nodes in the graph are concepts in the texts and edges in the graph are relations between concepts. Since AMR abstracts away from syntax and preserves only semantic information, it can be applied to many semantic related tasks such as summarization \cite{liu2015toward,liao2018abstract}, paraphrase detection \cite{issa2018abstract}, machine translation \cite{song2019semantic} and so on. 

Previous works on AMR parsing mainly focus on English, since AMR is designed for English texts and parallel corpus of non-English texts and AMRs are scarce. Early work of AMR announces that AMR is biased towards English and is not an interlingua \cite{banarescu2013abstract}. Besides, some studies show that aligning AMR with non-English language is not always possible \cite{xue2014not,hajic2014comparing}. However, recent studies \cite{damonte2018cross,blloshmi2020enabling} show that AMR parsers are able to recover AMR structures when there are structural differences between languages, which demonstrate that it is capable to overcome many translation divergences. Therefore, it is possible for us to parse texts in target (non-English) languages into AMRs. 

Another problem of cross-lingual AMR parsing is the scarcity of parallel corpus. Unlike machine translation or sentiment classification which have abundant resources on the Internet, we can only get non-English text and AMR pairs by human annotation. \citet{damonte2018cross} align a non-English token with an AMR node if they can be mapped to the same English token to construct training set. They further train a transition-based parser using the synthetic training set. They also attempt to translate test set into English and apply an English AMR parser. \citet{blloshmi2020enabling} build training data in two ways. One of the approaches is that they use gold parallel sentences and generate synthetic AMR annotations with the help of an English AMR parser. Another approach is to use gold English-AMR pairs and get non-English texts by a pre-trained machine translation system. They further use a sequence-to-graph parser \cite{zhang2019amr} to train a cross-lingual AMR parser. 

According to \cite{blloshmi2020enabling}, a cross-lingual AMR parser may predict the concepts less specific and accurate than the gold concepts. Therefore, we propose a new model introducing machine translation to enable our parser to predict more accurate concepts. In particular, we first build our training data similar to \cite{blloshmi2020enabling}, translating English texts into target languages. Our basic model is a sequence-to-sequence model, rather than the sequence-to-graph model used in \cite{blloshmi2020enabling}, since in English AMR parsing, sequence-to-sequence models can achieve state-of-the-art result with enough data for pre-training \cite{xu2020improving}. While training, we introduce bilingual input by concatenating translated target language texts and English texts as inputs. As for inference stage, the bilingual input is the concatenation of translated English texts and target language texts. We hope that our model can predict more accurate concepts with the help of the English tokens, while it can still preserve the meaning of the original texts if there are semantic shifts in the translated English texts. Besides, during training process, we also introduce an auxiliary task, requiring the decoder to restore English input tokens, which also aims at enhancing the ability of our parser to predict concepts. Our parser outperforms previous state-of-the-art parser XL-AMR \cite{blloshmi2020enabling} on LDC2020T07 dataset \cite{cai2013smatch} by about 10.6 points of Smatch F1 score on average, which demonstrates the efficacy of our proposed cross-lingual AMR parser. 

Our main contributions are summarized as follows:

\begin{itemize}
    \item We introduce bilingual inputs and an auxiliary task to a seq2seq cross-lingual AMR parser, aiming to enable the parser to make better use of bilingual information and predict more accurate concepts. 
    \item Our parser surpasses the best previously reported results of Smatch F1 score on LDC2020T07 by a large margin. The results demonstrate the effectiveness of our parser. Ablation studies show the usefulness of the model modules. Codes are public available \footnote{https://github.com/headacheboy/cross-lingual-amr-parsing}. 
    \item We further carry out experiments to investigate the influence of incorporating pre-training models into our cross-lingual AMR parser. 
\end{itemize}

\section{Related Work}

Abstract Meaning Representation (AMR) \cite{banarescu2013abstract} parsing is becoming popular recently. Some of previous works \cite{flanigan2014discriminative,lyu2018amr,zhang2019amr} solve this problem with a two-stage approach. They first project words in sentences to AMR concepts, followed by relation identification. Transition-based parsing is applied by \cite{wang2015transition,wang2015boosting,damonte2017incremental,liu2018amr,guo2018better,naseem2019rewarding,lee2020pushing}. They align words with AMR concepts and then take different actions based on different processed words to link edges or insert new nodes. Due to the recent development in sequence-to-sequence model, several works employ it to parse texts into AMRs \cite{konstas2017neural,van2017neural,ge2019modeling,xu2020improving}. They linearize AMR graphs and leverage character-level or word-level sequence-to-sequence model. Sequence-to-graph model is proposed to enable the decoder to better model the graph structure. \citet{zhang2019amr} first use a sequence-to-sequence model to predict concepts and use a biaffine classifier to predict edges. \citet{zhang2019broad} propose a one-stage sequence-to-graph model, predicting concepts and relations at the same time. \citet{cai_lam_2020_amr} regard AMR parsing as dual decisions on input sequences and constructing graphs. They therefore propose a sequence-to-graph method by first mapping an input words to a concept and then linking an edge based on the generated concepts. Recently, pre-training models have been proved to perform well in AMR parsing \cite{xu2020improving}. \citet{lee2020pushing} employ a self-training method to enhance a transition-based parser, which achieves the state-of-the-art Smatch F1 score in English AMR parsing. 

\citet{vanderwende2015amr} first carry out research of cross-lingual AMR parsing. They parse texts in target language into logical forms as a pivot, which are then parsed into AMR graphs. \citet{damonte2018cross} attempt to project non-English words to AMR concepts and use a transition-based parser to parse texts to AMR graphs. They also attempt to automatically translate non-English texts to English and exploit an English AMR parser. \citet{blloshmi2020enabling} try to generate synthetic training data by a machine translation system or an English AMR parser. They conduct experiments with a sequence-to-graph model in different settings, trying to find a best way to train with synthetic training data. Different from \cite{blloshmi2020enabling}, we treat cross-lingual AMR parsing as a sequence-to-sequence transduction problem and improve seq2seq models with bilingual input and auxiliary task.  

\section{Problem Setup}
\label{sec:problem setup}

\begin{figure}[tb]
    \centering
    \includegraphics[scale=0.5]{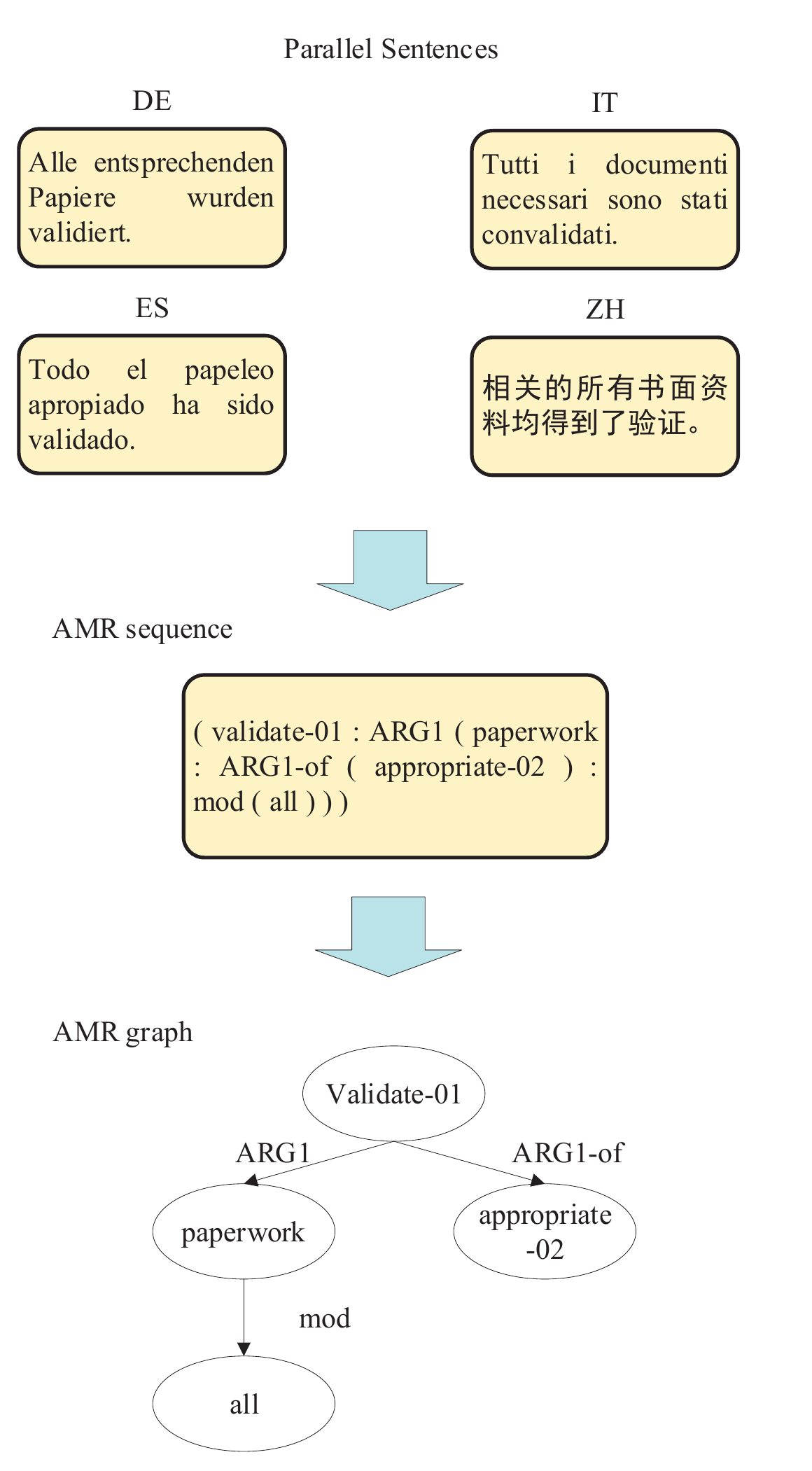}
    \caption{An example of cross-lingual AMR parsing. A Non-English text is first parsed into an AMR sequence and then the sequence is converted to an AMR graph.}
    \label{fig:example}
\end{figure}

Cross-lingual AMR parsing is the task of parsing non-English texts into AMR graphs corresponding to their English translation. In this task, nodes in AMR graphs are still English words, PropBank framesets or AMR keywords, which are the same as the original design of AMR. 

Figure \ref{fig:example} shows an example of cross-lingual AMR parsing. We define $X^{l}$ as an input sample in language \textit{l} and $X^{l}_i$ is the $i$-th token of it. $y$ is the corresponding AMR sequence derived from the AMR graph, and $y_i$ is the $i$-th token. The model should predict the AMR sequence $y$ first and then transform the sequence into a graph. 

\section{Our Proposed Model}

\begin{figure*}[ht]
    \centering
    \includegraphics[scale=0.5]{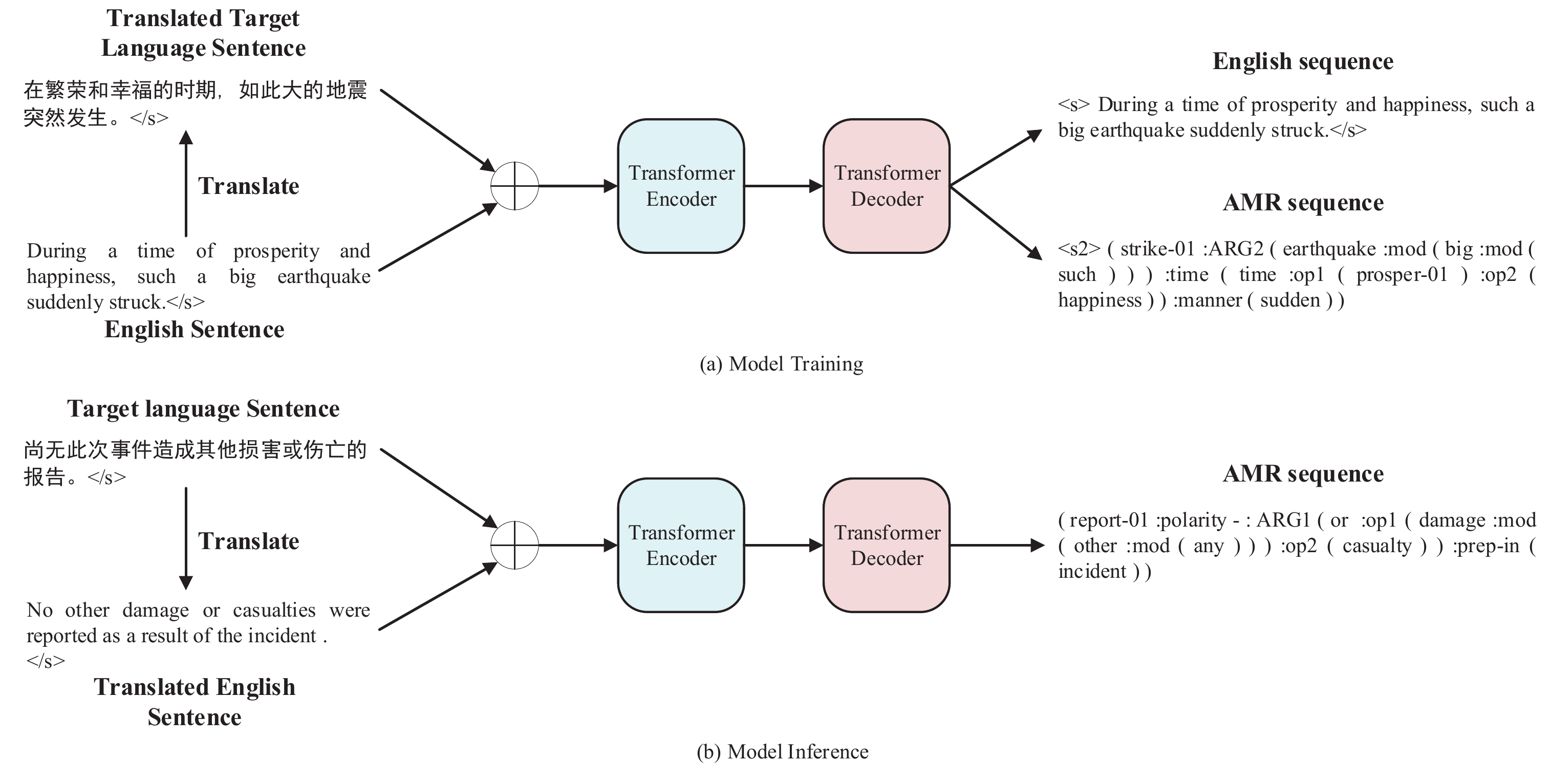}
    \caption{Overview of our proposed model. $\bigoplus$ is the concatenation operation. }
    \label{fig:model}
\end{figure*}

Figure \ref{fig:model} shows the training and inference processes of our proposed model. The basic model we adopt is Transformer \cite{vaswani2017attention} encoder-decoder model, since \citet{xu2020improving} show that it can achieves state-of-the-art result in English AMR parsing. We introduce the bilingual input to our model. When training the model, the bilingual input contains original English text and translated text in non-English target language, which may not be very accurate. During inference, the bilingual input is composed of translated English text and original text in target language. With the help of bilingual input, our model can better understand and preserve the semantics of target language texts, and predict more accurate concepts according to the translated English texts. Apart from predicting AMR sequences, the model is also required to predict the English input texts as an auxiliary objective, which can further help the model learn the exact meaning of input tokens and predict their corresponding concepts more accurately.

We will first introduce the way we obtain training data and the pre-processing and post-processing process in Section \ref{sec:synthetic_training_data} and Section \ref{sec:processing}, followed by introducing the basic sequence-to-sequence model, the bilingual input and the auxiliary task.

\subsection{Synthetic Training Data}
\label{sec:synthetic_training_data}

\citet{blloshmi2020enabling} propose two methods to generate parallel training data, namely \textbf{parallel sentences - silver AMR graphs} and \textbf{gold AMR graphs - silver translations}. The first approach means that we exploit human annotated parallel corpus of target languages and English and use an English parser to get the corresponding AMR graphs. The second approach means that we exploit human annotated English-AMR pairs and use a machine translation system to get texts in target languages. According to \cite{blloshmi2020enabling}, model training with data generated by gold English-AMR pairs performs better. We thus exploit this approach (i.e., \textbf{gold AMR graphs - silver translations}) to generate our data for training and validation. 

\subsection{Pre-Processing and Post-Processing}
\label{sec:processing}

Following \cite{van2017neural}, we first remove variables, since variables are only used to identify the same node in a graph and contain no semantic information, which may do harm to the model training process. We also remove wiki links (:wiki), since sequence-to-sequence model may link to non-existing objects of Wikipedia. As for co-referring nodes, we simply duplicate the concepts. It transforms an AMR graph into a tree. The final linearized AMR is the pre-order traversal of the tree. 

In post-processing, we should restore a predicted AMR sequence without variables, wiki links and co-referring nodes to a AMR graph. Following \cite{van2017neural}, We first restore variables and prune duplicated nodes, which brings co-reference back to the AMR sequence. \citet{van2017neural} use DBpedia Spotlight to restore wiki links. However, same entity in different language is linked to different pages in DBpedia, which makes it difficult for cross-lingual AMR parser to restore the wiki linking the entity in English. Different from \cite{van2017neural}, we restore a wiki link of a certain name if this name corresponds to the wiki link in training set. 

\subsection{Sequence-to-Sequence Model}

After pre-processing, both input texts and output AMRs are sequences. Hence we are able to apply a sequence-to-sequence model to accomplish cross-lingual AMR parsing. We use Transformer \cite{vaswani2017attention}, one of the most popular sequence-to-sequence model as our basic model. 

In order to be compatible with pre-training XLM-R \cite{conneau_etal_2020_unsupervised} model, the tokenizer and input vocabulary we used is the same as XLM-R. Subword unit such as byte pair encoding (BPE) \cite{sennrich2016neural} is commonly used to reduce the size of vocabulary. Thus, we exploit BPE to get our output vocabulary.

\begin{table}[tb]
    \small
    \centering
    \begin{tabular}{cccc}
    \toprule
       & Training  & Development & Test \\
       \midrule
       Number & 36521 & 1368 & 1371\\
        \bottomrule
    \end{tabular}
    \caption{Statistics of gold English-AMR dataset}
    \label{tab:table1}
\end{table}

\subsection{Bilingual Input}

AMR concepts heavily rely on the corresponding English texts. According to \cite{damonte2018cross}, a simple method that first translates the test set into English and then applies an English AMR parser can outperform their cross-lingual AMR parser. However, machine translation may introduce semantic shifts, which may do harm to the generation of AMRs. 

We therefore introduce the bilingual input. Since we do not have gold parallel corpus, we use machine translation to get the bilingual input. During training, we concatenate the translated text in target language mentioned in Section \ref{sec:synthetic_training_data} and the original English text as bilingual input. At the inference stage, we take the bilingual input by concatenating original target language texts and the translated English text. The model can better understand and preserve the semantic meanings of the input bilingual text. It can also predict more correct concepts, since the English tokens are also provided. 

\subsection{Auxiliary Task}

AMR concepts are composed of English words and Propbank frames. According to \cite{blloshmi2020enabling}, roughly 60\% of nodes in AMR 2.0 (LDC2017T10) are English words. What's more, Propbank predicates are similar to English words, such as predicate \textit{publish-01} and word \textit{publish}. We argue that if the decoder can restore the input tokens in English precisely, it can predict the corresponding concepts appropriately. 

We thus design an auxiliary task, requiring the decoder to predict the English input sequence. Inspired by multilingual machine translation \cite{johnson2017google}, we add a new BOS token indicating that the model should predict the English sequences instead of AMR sequences. The decoder predicting English sequences share the same weights as the decoder predicting AMR sequences. 

The final loss function is the weighted sum of loss functions of these two tasks: $Loss_{AMR}$ - the loss of AMR sequence prediction and $Loss_{Eng}$ - the loss of English sentence prediction. We adopt the cross-entropy loss for both tasks. 

\section{Implementation Details}

The coefficient of $Loss_{AMR}$ is 1, while the coefficient of $Loss_{Eng}$ is 0.5. We use Adam optimizer \cite{kingma2014adam} to optimize the final loss function. The number of transformer layers in both encoder and decoder is 6. The embedding size and hidden size are both 512 and the size of feed-forward network is 2048. The head number of multi-head attention is 8. We follow \cite{vaswani2017attention} to tune the learning rate each step and the warmup step is 4000. The learning rate for decoder at each step is half of this learning rate. Following \cite{blloshmi2020enabling}, we use machine translation system OPUS-MT \cite{tiedemann2020opus} to get our bilingual input. We use all data from different languages to train our model and the final model is able to parse sentences in different languages.

\section{Experiments}

\subsection{Dataset}

\begin{table}[tbp]
    \small
    \centering
    \begin{tabular}{cccccc}
        \toprule
         &  DE & IT & ES & ZH & AVG\\
         \midrule
         AMREager&  39.0 & 43.0 & 42.0 & 35.0 & 39.8 \\ 
         XL-AMR (Mul) & 49.9 & 53.5 & 53.2 & 41.0 & 49.4 \\
         XL-AMR (Mul*) & 52.1 & 56.7 & 56.2 & - & - \\
         XL-AMR (Lang) & 51.6 & 56.7 & 56.1 & 43.1 & 51.9 \\
         XL-AMR (Bi) & 53.0 & 58.1 & 58.0 & 41.5 & 52.7 \\
         \midrule
         Translate-test & 60.4 & 62.1 & 63.3 & 53.7 & 59.9 \\
         \midrule
         Ours & \textbf{64.0} & \textbf{65.4} & \textbf{67.3} & \textbf{56.5} & \textbf{63.3} \\
         \bottomrule
    \end{tabular}
    \caption{Smatch F1 scores of different models on German (DE), Italian (IT), Spanish (ES) and Chinese (ZH). (Mul) represents multilingual setting. (Mul*) represents multilingual setting except Chinese data. (Lang) represents language specific setting. (Bi) represents bilingual setting. The Smatch F1 score of English AMR parser is 68.3. Results of the best models are in bold.}
    \label{tab:main_results}
\end{table}

\begin{table*}[htb]
    \small
    \centering
    \begin{tabular}{ccccccccccccc}
        \toprule
          &  \multicolumn{4}{c}{AMREager} & \multicolumn{4}{c}{XL-AMR} & \multicolumn{4}{c}{Ours}\\
         \cmidrule(r){2-5} \cmidrule(r){6-9} \cmidrule{10-13} 
         Metric &  DE & IT & ES & ZH & DE & IT & ES & ZH & DE & IT & ES & ZH \\
         \cmidrule(r){1-1} \cmidrule(r){2-5} \cmidrule(r){6-9} \cmidrule{10-13} 
         Smatch & 39.1 & 43.2 & 42.1 & 34.6 & 53.0 & 58.1 & 58.0 & 43.1 & \textbf{64.0} & \textbf{65.4} & \textbf{67.3} & \textbf{56.5} \\
         \cmidrule(r){1-1} \cmidrule(r){2-5} \cmidrule(r){6-9} \cmidrule{10-13} 
         Unlabeled & 45.0 & 48.5 & 46.6 & 41.1 & 57.7 & 63.4 & 63.0 & 48.9 & \textbf{68.1} & \textbf{69.6} & \textbf{71.2} & \textbf{61.0} \\
         No WSD & 39.2 & 42.5 & 42.2 & 34.7 & 53.2 & 58.4 & 58.4 & 43.2 & \textbf{64.4} & \textbf{65.9} & \textbf{67.8} & \textbf{56.7} \\
         Reentrancies & 18.6 & 25.7 & 27.2 & 15.9 & 39.9 & 46.1 & 46.6 & 34.7 & \textbf{47.9} & \textbf{49.3} & \textbf{51.3} & \textbf{41.4}\\
         Concepts & 44.9 & 52.3 & 53.3 & 39.9 & 58.0 & 64.7 & 65.9 & 48.0 & \textbf{69.3} & \textbf{72.1} & \textbf{75.0} & \textbf{61.3}\\
         Named Ent. & 63.1 & 67.7 & 65.7 & 67.9 & 66.0 & 70.0 & 66.2 & 60.6 & \textbf{79.3} & \textbf{79.5} & \textbf{80.2} & \textbf{76.2}\\
         Wikification & 49.9 & 50.6 & 44.5 & 46.8 & 60.9 & 67.0 & 63.1 & 54.5 & \textbf{74.0} & \textbf{74.9} & \textbf{73.9} & \textbf{68.1} \\
         Negation & 18.6 & 22.3 & 19.8 & 6.8 & 11.7 & 29.2 & 23.4 & 12.8 & \textbf{47.1} & \textbf{52.6} & \textbf{55.6} & \textbf{36.6} \\
         SRL & 29.4 & 34.3 & 35.9 & 27.2 & 47.9 & 54.7 & 55.2 & 41.3 & \textbf{57.3} & \textbf{60.1} & \textbf{62.1} & \textbf{50.7} \\
         \bottomrule
    \end{tabular}
    \caption{Fine-grained results of different models on DE, IT, ES and ZH. Best results are in bold.}
    \label{tab:finegrain}
\end{table*}

The released test set, LDC2020T07, contains four translations of test set of AMR 2.0, including German (DE), Italian (IT), Spanish (ES) and Chinese(ZH). 

As is mentioned in Section \ref{sec:synthetic_training_data}, we translate the sentences in a gold English-AMR dataset to get training and development data with OPUS-MT. We use AMR 2.0 as our gold English-AMR dataset. We also translate test sets in German, Italian, Spanish and Chinese back to English as input texts. The statistics of AMR 2.0 are shown in Table \ref{tab:table1}. 

\subsection{Evaluation Metric}

Smatch \cite{cai2013smatch} is the evaluation metric of AMR parsing. In this evaluation metric, AMR graph is regarded as several triples. Smatch counts the numbers of matched triples and outputs the score based on total numbers of triples of two AMR graphs. We use the Smatch scripts available online \footnote{https://github.com/sheng-z/stog}. 

Following \cite{damonte2017incremental}, we also introduce many fine-grained evaluations in order to evaluate the quality of the predicted AMR graphs in different aspects. We omit the details of these fine-grained evaluations here, which can be found in \cite{damonte2017incremental}. 

\subsection{Main Results}

\begin{table*}[tb]
    \small
    \centering
    \begin{tabular}{cccccccc}
        \toprule
        & S2S & \multicolumn{2}{c}{S2S + Bilingual Input} & \multicolumn{2}{c}{S2S + Auxiliary} & \multicolumn{2}{c}{Full Model} \\
        \cmidrule(r){2-2} \cmidrule(r){3-4} \cmidrule(r){5-6} \cmidrule(r){7-8}
        & F1 & F1 & $\Delta$ & F1 & $\Delta$ & F1 & $\Delta$ \\
        \cmidrule(r){1-2} \cmidrule(r){3-4} \cmidrule(r){5-6} \cmidrule(r){7-8}
        Smatch & 53.1 & 57.5 & 4.4 & 58.6 & 5.5 & 63.3 & 10.2\\
        \cmidrule(r){1-2} \cmidrule(r){3-4} \cmidrule(r){5-6} \cmidrule(r){7-8}
        Unlabeled & 57.7 & 59.2 & 1.5 & 58.9 & 1.2 & 67.5 & 9.8\\
        No WSD & 53.4 & 59.2 & 5.8 & 58.9 &5.5 & 63.7 & 10.3\\
        Reentrancies & 38.4 & 41.9 & 3.5 & 42.7 & 4.3 & 47.5 & 9.1 \\
        Concepts & 57.3 & 66.4 & 9.1 & 62.7 & 5.4 & 69.4 & 12.1 \\
        Named Ent. & 73.7 & 75.0 & 1.3 & 77.2 & 3.5 & 78.8 & 5.1\\
        Wiki & 62.1 & 69.0 & 6.9 & 69.2 & 7.1 & 72.7 & 10.6\\
        Negation & 32.9 & 44.1 & 11.2 & 39.5 & 6.6 & 48.0 & 15.1\\
        SRL & 47.4 & 52.1 & 4.7 & 52.2 & 4.8 & 57.6 & 10.2\\
    \bottomrule
    \end{tabular}
    \caption{Smatch and fine-grained results of ablation study. The listed Scores are avereage F1 score for different metrics on four test sets. $\Delta$ represents the model improvement compared with basic sequence-to-sequence(s2s) model.}
    \label{tab:ablation study}
\end{table*}

We compare our model with previous works and baseline methods including:

\begin{itemize}
    \item \textbf{AMREager}. This is the model proposed by \citet{damonte2018cross}. They assume that if a word $X^{l}_t$ in target language is aligned with the word $X^{en}_u$ in English and the English word aligns with AMR concept $y_i$, $X^{l}_t$ can be aligned with $y_i$. Based on this assumption, they project AMR annotations to target languages and further train a transition-based AMR parser \cite{damonte2017incremental} as in English. 
    \item \textbf{XL-AMR}. This is the model proposed by \cite{blloshmi2020enabling}. When conducting experiments of their best model, they first generate synthetic training and validation data by machine translation. They then train an AMR parser on target language with a sequence-to-graph parser. They experiment XL-AMR with many different settings: language specific setting, bilingual setting and multilingual setting. Language specific setting means that they only use target language data to train the model. Bilingual setting represents training with target language data and English data. Multilingual setting represents training with data in all languages. They also experiment multilingual setting except Chinese data because they found training with Chinese data will lower the results.
    \item \textbf{Translate-test}. This method first translates target language texts into English and uses an English AMR parser to predict the final AMR graphs. For fair comparison, we choose the sequence-to-sequence model as the English AMR parser. The encoder, decoder and hyper-parameters of these modules are the same as those in our model. We use only English texts as input and do not apply the auxiliary task in the training of English parser. Note that this baseline is not compared in \cite{blloshmi2020enabling} and we show it is a very strong baseline. 
\end{itemize}

\begin{figure}[htb]
    \centering
    \includegraphics[scale=0.45]{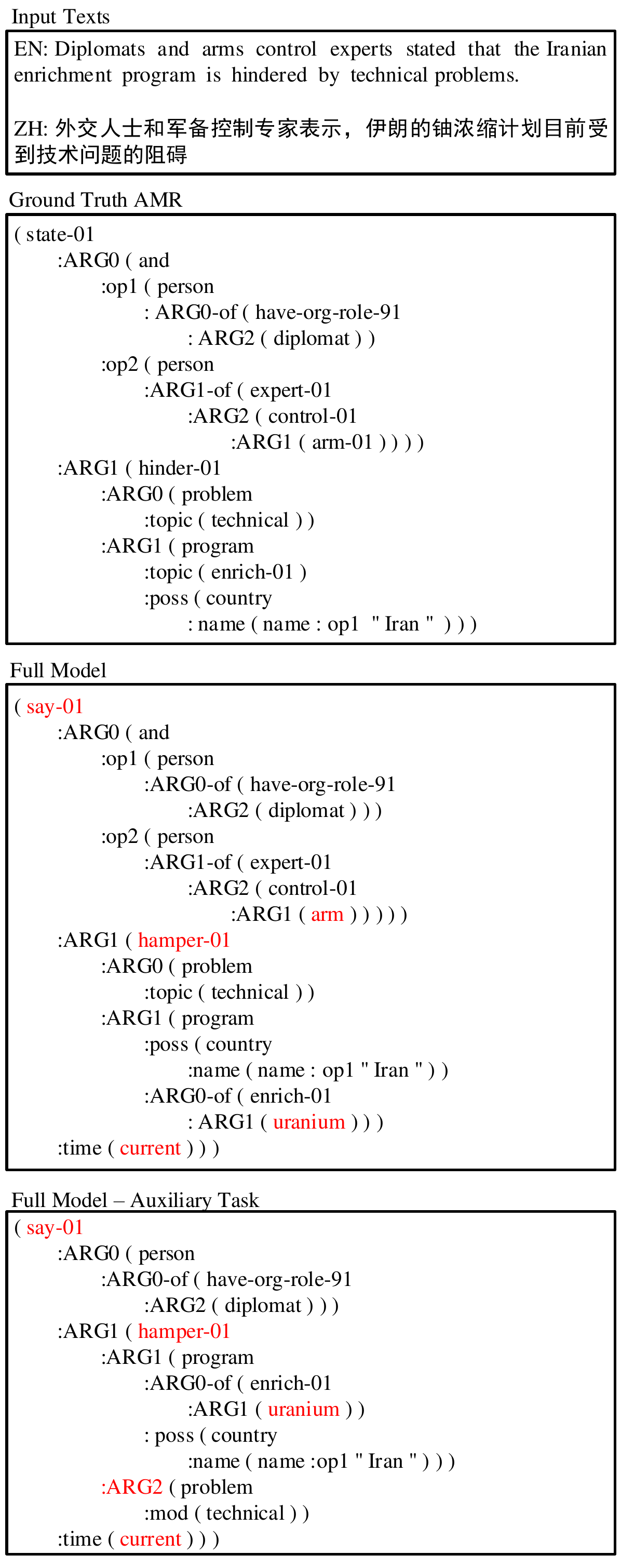}
    \caption{An example of AMR predicted by models with and without auxiliary task. We mark the error concepts and relations in red. }
    \label{fig:case_study_2}
\end{figure}

The comparison results are shown in Table \ref{tab:main_results}. Our model outperforms previous best model XL-AMR in different settings by a large margin. As for languages that share similarity with English, namely German, Italian, Spanish, our proposed model achieves substantial improvement on Smatch F1 score by about 10 points. When it comes to languages that has linguistic differences with English, namely Chinese, our model performs better, surpassing XL-AMR by 13.4 points on Smatch F1 score. The Translate-test method is a strong baseline because of the quality of machine translation. It outperforms previous reported results by a large margin, which reveals that English information is significantly beneficial to AMR prediction. In this work, our model also surpasses this method by 3.4 points and achieves the new state-of-the-art results. 

Table \ref{tab:finegrain} lists the fine-grained evaluation results of AMREager, the best XL-AMR model and our model. Our proposed model achieves substantially higher performance by about 10 points for each fine-grained task except \textit{Negation}. As for \textit{Negation}, our model achieves over 20 points higher than XL-AMR. These results demonstrates that our model not only predicts better concepts but also predicts better relations between concepts.

\subsection{Ablation Study}

In order to verify the effectiveness of the bilingual input and the auxiliary task in our model, we carry out several ablation experiments.  

Table \ref{tab:ablation study} shows the Smatch score and fine-grained results of ablation study. Compared with the basic sequence-to-sequence model, the bilingual input can improve the Smatch F1 score by 4.4 points on average. The introduction of auxiliary task brings 5.5 points improvement of Smatch on average. Our full model makes use of both bilingual input and auxiliary task at the same time, improving Smatch scores by 10.2 points, which indicates that each module is very beneficial to the performance of our model. 

Fine-grained results further demonstrate the effectiveness of our modules. As Table \ref{tab:ablation study} shows, F1 score for \textit{Concepts} improves substantially, which demonstrates that our proposed modules can actually help the parser predict more accurate concepts. Besides, fine-grained evaluation \textit{Negation} achieves the highest improvement, revealing that our proposed modules enable the parser to understand the semantics of the input texts better. 

\subsection{Effect of Pre-trained Models on Cross-Lingual AMR Parsing}

\begin{figure*}
    \centering
    \includegraphics[scale=0.45]{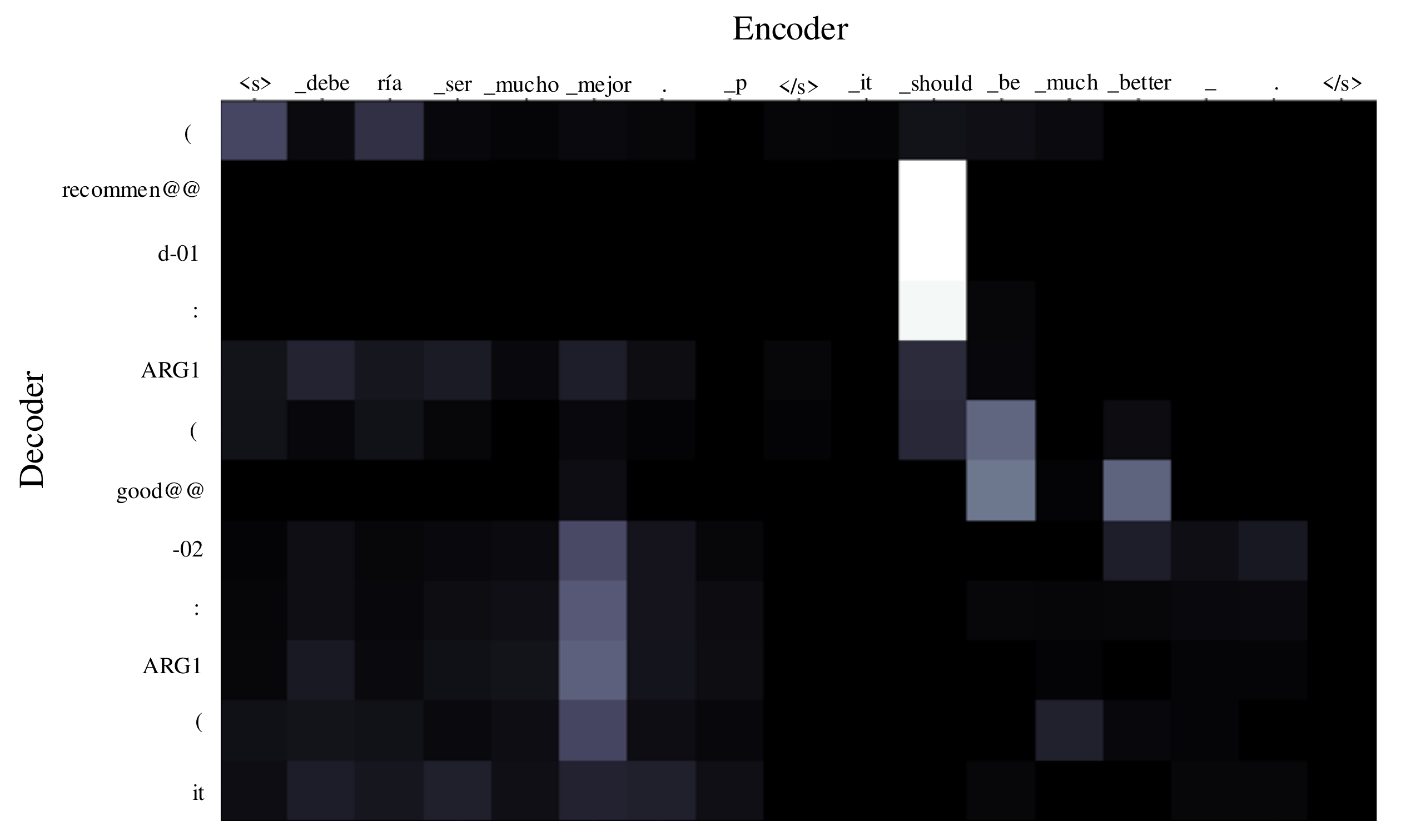}
    \caption{An attention heatmap from one head in the second decoder layer. The attention weight is higher if the color is lighter.}
    \label{fig:attention_map}
\end{figure*}

\begin{table}[tb]
    \small
    \centering
    \begin{tabular}{cccccc}
    \toprule
        Model & DE & IT & ES & ZH & AVG \\
         \midrule
         Full Model & 64.0 & 65.4 & 67.3 & 56.5 & 63.3 \\
         \midrule
         + XLM-R & 66.1 & 67.9 & 69.6 & 57.9 & 65.4 \\
         + dec & 64.9 & 66.7 & 68.5 & 57.4 & 64.4 \\
         + XLM-R \& dec & 68.3 & 70.0 & 71.9 & 59.6 & 67.5\\
    \bottomrule
    \end{tabular}
    \caption{Smatch scores of models employing pre-trained models.  }
    \label{tab:pre_train}
\end{table}

\begin{figure}[tb]
    \centering
    \includegraphics[scale=0.5]{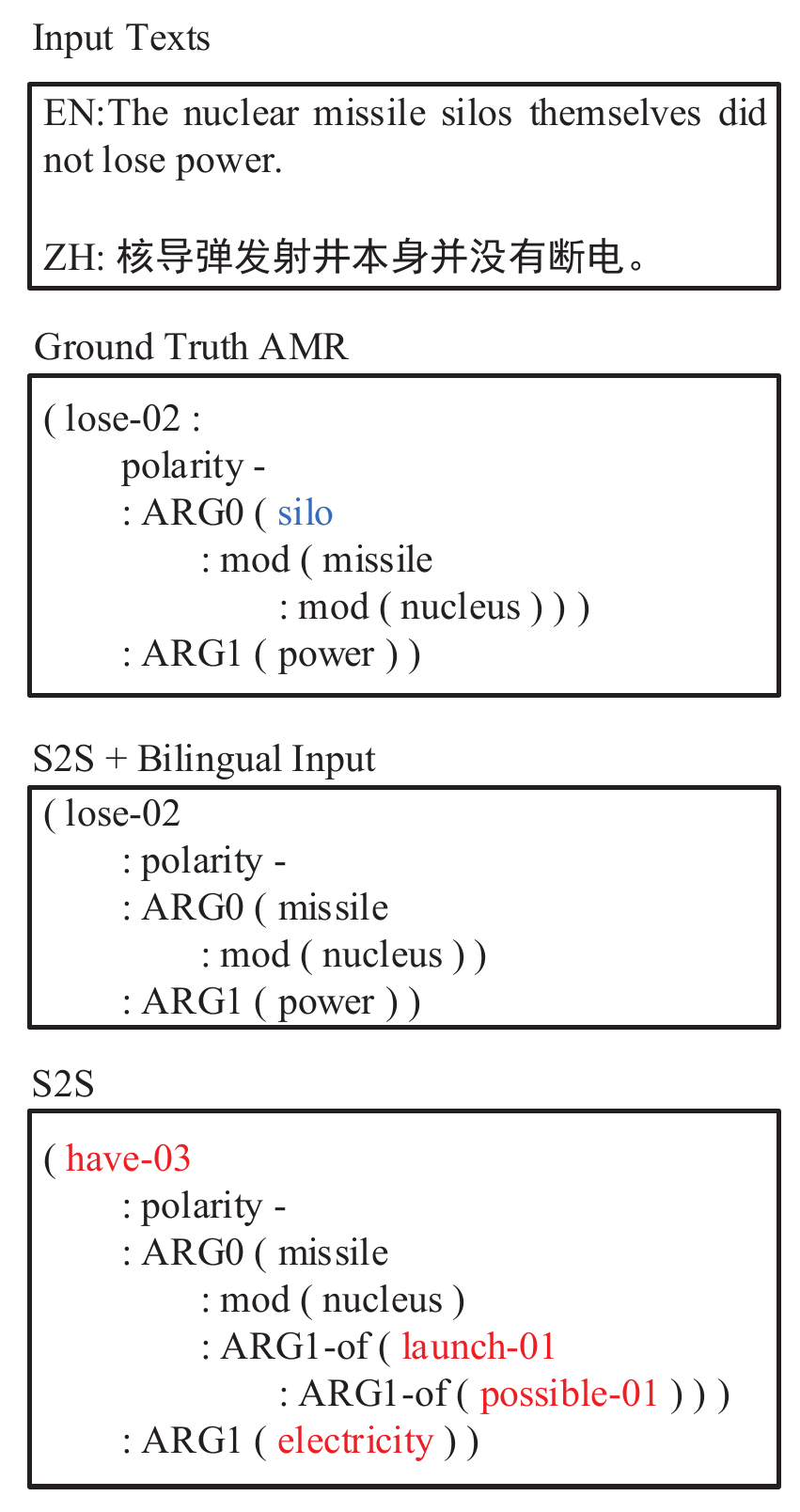}
    \caption{An example of AMR predicted by S2S and S2S + Bilingual Input. We mark the missing concept in blue and mark the inaccurate concepts in red. }
    \label{fig:case_study_1}
\end{figure}

Recently, pre-training models on cross-lingual tasks have been proposed. Pre-training models, such as mBert \cite{devlin2019bert}, XLM \cite{conneau2019cross}, XLM-R \cite{conneau2020unsupervised} and mBart \cite{liu2020multilingual} achieve state-of-the-art results on many tasks such as machine translation, cross-lingual natural language inference and so on. In our experiment, we exploit XLM-R \cite{conneau2020unsupervised} as the input embeddings of the model. 

When training an English AMR parser, \citet{xu2020improving} first pre-train the model on large scale synthetic data and fine-tune it on gold English-AMR data. Since cross-lingual AMR parsing shares the same output formats with AMR parsing, we can employ the decoder of \cite{xu2020improving} to initialize our decoder and further finetune the cross-lingual AMR parser. 

Results are listed in Table \ref{tab:pre_train}. The performance of our parser with XLM-R embedding improves by 2.1 points on Smatch score, while our parser finetuning pre-trained AMR decoder achieves 1.1 points improvement. We further employ both XLM-R embedding and pre-trained AMR decoder and the average Smatch score is 67.5. The results show that pre-trained cross-lingual embeddings like XLM-R as well as the pre-trained decoder can help the parser predict better AMR graphs. 

\section{Analysis}

Figure \ref{fig:case_study_2} shows several AMRs parsed by models with and without auxiliary task. The AMR predicted by model without auxiliary task misses many concepts, while our full model predicts them correctly. What's more, our full model can predict relations of concepts more accurately as well. For example, the full model adds \textit{ARG0} between \textit{hamper-01} and \textit{problem}, retaining semantic information of the original sentence. The model trained without auxiliary task predicts the relation \textit{ARG2} instead, changing the meaning of the original sentence. 

Another example in Figure \ref{fig:case_study_1} shows the efficacy of bilingual inputs. The AMR parsed by basic sequence-to-sequence model does not contain correct semantics. This model predicts many erroneous concepts such as \textit{launch-01}, \textit{possible-01}. Besides, the semantics of original sentence \textit{did not lose power} is changed into \textit{have no electricity}. The AMR produced by sequence-to-sequence model with bilingual input is almost correct except missing of concept \textit{silo}. This example reveals that our bilingual input enables the parser to predict more accurate concepts and preserve the semantics of the sentence. 

We also show an attention heatmap of an example in test set in Figure \ref{fig:attention_map}. This attention pattern shows that our parser can predict AMR tokens based on English translation (e.g. \textit{recommend-01}) and based on both English and Spanish tokens (e.g. \textit{good-02}). 

\section{Conclusion}

In this paper, we focus on cross-lingual AMR parsing. Previous works have deficiency in predicting correct AMR concepts. We thus introduce bilingual inputs as well as an auxiliary task to predict more accurate concepts and their relations in AMR graphs. Empirical results on data in German, Italian, Spanish and Chinese demonstrate the efficacy of our proposed method. We also conduct ablation study to further verify the significance of the bilingual inputs and auxiliary task. For future work, we will attempt to adapt other methods used in English AMR parsing to cross-lingual AMR parsing, such as pre-training and self-training. 

\section*{Acknowledgments}

This work was supported by National Natural Science Foundation of China (61772036), Beijing Academy of Artificial Intelligence (BAAI) and Key Laboratory of Science, Technology and Standard in Press Industry (Key Laboratory of Intelligent Press Media Technology). We appreciate the anonymous reviewers for their helpful comments. Xiaojun Wan is the corresponding author.

\bibliographystyle{acl_natbib}
\bibliography{anthology,acl2021}

\begin{thebibliography}{37}
\expandafter\ifx\csname natexlab\endcsname\relax\def\natexlab#1{#1}\fi

\bibitem[{Banarescu et~al.(2013)Banarescu, Bonial, Cai, Georgescu, Griffitt,
  Hermjakob, Knight, Koehn, Palmer, and Schneider}]{banarescu2013abstract}
Laura Banarescu, Claire Bonial, Shu Cai, Madalina Georgescu, Kira Griffitt, Ulf
  Hermjakob, Kevin Knight, Philipp Koehn, Martha Palmer, and Nathan Schneider.
  2013.
\newblock Abstract meaning representation for sembanking.
\newblock In \emph{Proceedings of the 7th linguistic annotation workshop and
  interoperability with discourse}, pages 178--186.

\bibitem[{Blloshmi et~al.(2020)Blloshmi, Tripodi, and
  Navigli}]{blloshmi2020enabling}
Rexhina Blloshmi, Rocco Tripodi, and Roberto Navigli. 2020.
\newblock Enabling cross-lingual amr parsing with transfer learning techniques.
\newblock In \emph{Proceedings of the 2020 Conference on Empirical Methods in
  Natural Language Processing (EMNLP)}, pages 2487--2500.

\bibitem[{Cai and Lam(2020)}]{cai_lam_2020_amr}
Deng Cai and Wai Lam. 2020.
\newblock \href {https://doi.org/10.18653/v1/2020.acl-main.119} {{AMR} parsing
  via graph-sequence iterative inference}.
\newblock In \emph{Proceedings of the 58th Annual Meeting of the Association
  for Computational Linguistics}, pages 1290--1301, Online. Association for
  Computational Linguistics.

\bibitem[{Cai and Knight(2013)}]{cai2013smatch}
Shu Cai and Kevin Knight. 2013.
\newblock Smatch: an evaluation metric for semantic feature structures.
\newblock In \emph{Proceedings of the 51st Annual Meeting of the Association
  for Computational Linguistics (Volume 2: Short Papers)}, pages 748--752.

\bibitem[{Conneau et~al.(2020{\natexlab{a}})Conneau, Khandelwal, Goyal,
  Chaudhary, Wenzek, Guzm{\'a}n, Grave, Ott, Zettlemoyer, and
  Stoyanov}]{conneau_etal_2020_unsupervised}
Alexis Conneau, Kartikay Khandelwal, Naman Goyal, Vishrav Chaudhary, Guillaume
  Wenzek, Francisco Guzm{\'a}n, Edouard Grave, Myle Ott, Luke Zettlemoyer, and
  Veselin Stoyanov. 2020{\natexlab{a}}.
\newblock \href {https://doi.org/10.18653/v1/2020.acl-main.747} {Unsupervised
  cross-lingual representation learning at scale}.
\newblock In \emph{Proceedings of the 58th Annual Meeting of the Association
  for Computational Linguistics}, pages 8440--8451, Online. Association for
  Computational Linguistics.

\bibitem[{Conneau et~al.(2020{\natexlab{b}})Conneau, Khandelwal, Goyal,
  Chaudhary, Wenzek, Guzm{\'a}n, Grave, Ott, Zettlemoyer, and
  Stoyanov}]{conneau2020unsupervised}
Alexis Conneau, Kartikay Khandelwal, Naman Goyal, Vishrav Chaudhary, Guillaume
  Wenzek, Francisco Guzm{\'a}n, {\'E}douard Grave, Myle Ott, Luke Zettlemoyer,
  and Veselin Stoyanov. 2020{\natexlab{b}}.
\newblock Unsupervised cross-lingual representation learning at scale.
\newblock In \emph{Proceedings of the 58th Annual Meeting of the Association
  for Computational Linguistics}, pages 8440--8451.

\bibitem[{Conneau and Lample(2019)}]{conneau2019cross}
Alexis Conneau and Guillaume Lample. 2019.
\newblock Cross-lingual language model pretraining.
\newblock In \emph{Advances in Neural Information Processing Systems}, pages
  7059--7069.

\bibitem[{Damonte and Cohen(2018)}]{damonte2018cross}
Marco Damonte and Shay~B Cohen. 2018.
\newblock Cross-lingual abstract meaning representation parsing.
\newblock In \emph{Proceedings of the 2018 Conference of the North American
  Chapter of the Association for Computational Linguistics: Human Language
  Technologies, Volume 1 (Long Papers)}, pages 1146--1155.

\bibitem[{Damonte et~al.(2017)Damonte, Cohen, and
  Satta}]{damonte2017incremental}
Marco Damonte, Shay~B Cohen, and Giorgio Satta. 2017.
\newblock An incremental parser for abstract meaning representation.
\newblock In \emph{Proceedings of the 15th Conference of the European Chapter
  of the Association for Computational Linguistics: Volume 1, Long Papers},
  pages 536--546.

\bibitem[{Devlin et~al.(2019)Devlin, Chang, Lee, and
  Toutanova}]{devlin2019bert}
Jacob Devlin, Ming-Wei Chang, Kenton Lee, and Kristina Toutanova. 2019.
\newblock Bert: Pre-training of deep bidirectional transformers for language
  understanding.
\newblock In \emph{Proceedings of the 2019 Conference of the North American
  Chapter of the Association for Computational Linguistics: Human Language
  Technologies, Volume 1 (Long and Short Papers)}, pages 4171--4186.

\bibitem[{Flanigan et~al.(2014)Flanigan, Thomson, Carbonell, Dyer, and
  Smith}]{flanigan2014discriminative}
Jeffrey Flanigan, Sam Thomson, Jaime~G Carbonell, Chris Dyer, and Noah~A Smith.
  2014.
\newblock A discriminative graph-based parser for the abstract meaning
  representation.
\newblock In \emph{Proceedings of the 52nd Annual Meeting of the Association
  for Computational Linguistics (Volume 1: Long Papers)}, pages 1426--1436.

\bibitem[{Ge et~al.(2019)Ge, Li, Zhu, and Li}]{ge2019modeling}
Donglai Ge, Junhui Li, Muhua Zhu, and Shoushan Li. 2019.
\newblock Modeling source syntax and semantics for neural amr parsing.
\newblock In \emph{IJCAI}, pages 4975--4981.

\bibitem[{Guo and Lu(2018)}]{guo2018better}
Zhijiang Guo and Wei Lu. 2018.
\newblock Better transition-based amr parsing with a refined search space.
\newblock In \emph{Proceedings of the 2018 Conference on Empirical Methods in
  Natural Language Processing}, pages 1712--1722.

\bibitem[{Hajic et~al.(2014)Hajic, Bojar, and Uresova}]{hajic2014comparing}
Jan Hajic, Ond{\v{r}}ej Bojar, and Zdenka Uresova. 2014.
\newblock Comparing czech and english amrs.
\newblock In \emph{Proceedings of Workshop on Lexical and Grammatical Resources
  for Language Processing}, pages 55--64.

\bibitem[{Issa et~al.(2018)Issa, Damonte, Cohen, Yan, and
  Chang}]{issa2018abstract}
Fuad Issa, Marco Damonte, Shay~B Cohen, Xiaohui Yan, and Yi~Chang. 2018.
\newblock Abstract meaning representation for paraphrase detection.
\newblock In \emph{Proceedings of the 2018 Conference of the North American
  Chapter of the Association for Computational Linguistics: Human Language
  Technologies, Volume 1 (Long Papers)}, pages 442--452.

\bibitem[{Johnson et~al.(2017)Johnson, Schuster, Le, Krikun, Wu, Chen, Thorat,
  Vi{\'e}gas, Wattenberg, Corrado et~al.}]{johnson2017google}
Melvin Johnson, Mike Schuster, Quoc~V Le, Maxim Krikun, Yonghui Wu, Zhifeng
  Chen, Nikhil Thorat, Fernanda Vi{\'e}gas, Martin Wattenberg, Greg Corrado,
  et~al. 2017.
\newblock Google’s multilingual neural machine translation system: Enabling
  zero-shot translation.
\newblock \emph{Transactions of the Association for Computational Linguistics},
  5:339--351.

\bibitem[{Kingma and Ba(2015)}]{kingma2014adam}
Diederik~P. Kingma and Jimmy Ba. 2015.
\newblock \href {http://arxiv.org/abs/1412.6980} {Adam: {A} method for
  stochastic optimization}.
\newblock In \emph{3rd International Conference on Learning Representations,
  {ICLR} 2015, San Diego, CA, USA, May 7-9, 2015, Conference Track
  Proceedings}.

\bibitem[{Konstas et~al.(2017)Konstas, Iyer, Yatskar, Choi, and
  Zettlemoyer}]{konstas2017neural}
Ioannis Konstas, Srinivasan Iyer, Mark Yatskar, Yejin Choi, and Luke
  Zettlemoyer. 2017.
\newblock Neural amr: Sequence-to-sequence models for parsing and generation.
\newblock In \emph{Proceedings of the 55th Annual Meeting of the Association
  for Computational Linguistics (Volume 1: Long Papers)}, pages 146--157.

\bibitem[{Lee et~al.(2020)Lee, Astudillo, Naseem, Reddy, Florian, and
  Roukos}]{lee2020pushing}
Young-Suk Lee, Ram{\'o}n~Fernandez Astudillo, Tahira Naseem, Revanth~Gangi
  Reddy, Radu Florian, and Salim Roukos. 2020.
\newblock Pushing the limits of amr parsing with self-learning.
\newblock In \emph{Proceedings of the 2020 Conference on Empirical Methods in
  Natural Language Processing: Findings}, pages 3208--3214.

\bibitem[{Liao et~al.(2018)Liao, Lebanoff, and Liu}]{liao2018abstract}
Kexin Liao, Logan Lebanoff, and Fei Liu. 2018.
\newblock Abstract meaning representation for multi-document summarization.
\newblock In \emph{Proceedings of the 27th International Conference on
  Computational Linguistics}, pages 1178--1190.

\bibitem[{Liu et~al.(2015)Liu, Flanigan, Thomson, Sadeh, and
  Smith}]{liu2015toward}
Fei Liu, Jeffrey Flanigan, Sam Thomson, Norman Sadeh, and Noah~A Smith. 2015.
\newblock Toward abstractive summarization using semantic representations.
\newblock In \emph{Proceedings of the 2015 Conference of the North American
  Chapter of the Association for Computational Linguistics: Human Language
  Technologies}, pages 1077--1086.

\bibitem[{Liu et~al.(2018)Liu, Che, Zheng, Qin, and Liu}]{liu2018amr}
Yijia Liu, Wanxiang Che, Bo~Zheng, Bing Qin, and Ting Liu. 2018.
\newblock An amr aligner tuned by transition-based parser.
\newblock In \emph{Proceedings of the 2018 Conference on Empirical Methods in
  Natural Language Processing}, pages 2422--2430.

\bibitem[{Liu et~al.(2020)Liu, Gu, Goyal, Li, Edunov, Ghazvininejad, Lewis, and
  Zettlemoyer}]{liu2020multilingual}
Yinhan Liu, Jiatao Gu, Naman Goyal, Xian Li, Sergey Edunov, Marjan
  Ghazvininejad, Mike Lewis, and Luke Zettlemoyer. 2020.
\newblock Multilingual denoising pre-training for neural machine translation.
\newblock \emph{Transactions of the Association for Computational Linguistics},
  8:726--742.

\bibitem[{Lyu and Titov(2018)}]{lyu2018amr}
Chunchuan Lyu and Ivan Titov. 2018.
\newblock Amr parsing as graph prediction with latent alignment.
\newblock In \emph{Proceedings of the 56th Annual Meeting of the Association
  for Computational Linguistics (Volume 1: Long Papers)}, pages 397--407.

\bibitem[{Naseem et~al.(2019)Naseem, Shah, Wan, Florian, Roukos, and
  Ballesteros}]{naseem2019rewarding}
Tahira Naseem, Abhishek Shah, Hui Wan, Radu Florian, Salim Roukos, and Miguel
  Ballesteros. 2019.
\newblock Rewarding smatch: Transition-based amr parsing with reinforcement
  learning.
\newblock In \emph{Proceedings of the 57th Annual Meeting of the Association
  for Computational Linguistics}, pages 4586--4592.

\bibitem[{van Noord and Bos(2017)}]{van2017neural}
Rik van Noord and Johan Bos. 2017.
\newblock Neural semantic parsing by character-based translation: Experiments
  with abstract meaning representations.
\newblock \emph{Computational Linguistics in the Netherlands Journal},
  7:93--108.

\bibitem[{Sennrich et~al.(2016)Sennrich, Haddow, and
  Birch}]{sennrich2016neural}
Rico Sennrich, Barry Haddow, and Alexandra Birch. 2016.
\newblock Neural machine translation of rare words with subword units.
\newblock In \emph{Proceedings of the 54th Annual Meeting of the Association
  for Computational Linguistics (Volume 1: Long Papers)}, pages 1715--1725.

\bibitem[{Song et~al.(2019)Song, Gildea, Zhang, Wang, and
  Su}]{song2019semantic}
Linfeng Song, Daniel Gildea, Yue Zhang, Zhiguo Wang, and Jinsong Su. 2019.
\newblock Semantic neural machine translation using amr.
\newblock \emph{Transactions of the Association for Computational Linguistics},
  7:19--31.

\bibitem[{Tiedemann and Thottingal(2020)}]{tiedemann2020opus}
J{\"o}rg Tiedemann and Santhosh Thottingal. 2020.
\newblock Opus-mt--building open translation services for the world.
\newblock In \emph{Proceedings of the 22nd Annual Conference of the European
  Association for Machine Translation}, pages 479--480.

\bibitem[{Vanderwende et~al.(2015)Vanderwende, Menezes, and
  Quirk}]{vanderwende2015amr}
Lucy Vanderwende, Arul Menezes, and Chris Quirk. 2015.
\newblock An amr parser for english, french, german, spanish and japanese and a
  new amr-annotated corpus.
\newblock In \emph{Proceedings of the 2015 Conference of the North American
  Chapter of the Association for Computational Linguistics: Demonstrations},
  pages 26--30.

\bibitem[{Vaswani et~al.(2017)Vaswani, Shazeer, Parmar, Uszkoreit, Jones,
  Gomez, Kaiser, and Polosukhin}]{vaswani2017attention}
Ashish Vaswani, Noam Shazeer, Niki Parmar, Jakob Uszkoreit, Llion Jones,
  Aidan~N Gomez, {\L}ukasz Kaiser, and Illia Polosukhin. 2017.
\newblock Attention is all you need.
\newblock In \emph{Advances in neural information processing systems}, pages
  5998--6008.

\bibitem[{Wang et~al.(2015{\natexlab{a}})Wang, Xue, and
  Pradhan}]{wang2015boosting}
Chuan Wang, Nianwen Xue, and Sameer Pradhan. 2015{\natexlab{a}}.
\newblock Boosting transition-based amr parsing with refined actions and
  auxiliary analyzers.
\newblock In \emph{Proceedings of the 53rd Annual Meeting of the Association
  for Computational Linguistics and the 7th International Joint Conference on
  Natural Language Processing (Volume 2: Short Papers)}, pages 857--862.

\bibitem[{Wang et~al.(2015{\natexlab{b}})Wang, Xue, and
  Pradhan}]{wang2015transition}
Chuan Wang, Nianwen Xue, and Sameer Pradhan. 2015{\natexlab{b}}.
\newblock A transition-based algorithm for amr parsing.
\newblock In \emph{Proceedings of the 2015 Conference of the North American
  Chapter of the Association for Computational Linguistics: Human Language
  Technologies}, pages 366--375.

\bibitem[{Xu et~al.(2020)Xu, Li, Zhu, Zhang, and Zhou}]{xu2020improving}
Dongqin Xu, Junhui Li, Muhua Zhu, Min Zhang, and Guodong Zhou. 2020.
\newblock Improving amr parsing with sequence-to-sequence pre-training.
\newblock In \emph{Proceedings of the 2020 Conference on Empirical Methods in
  Natural Language Processing (EMNLP)}, pages 2501--2511.

\bibitem[{Xue et~al.(2014)Xue, Bojar, Hajic, Palmer, Uresova, and
  Zhang}]{xue2014not}
Nianwen Xue, Ondrej Bojar, Jan Hajic, Martha Palmer, Zdenka Uresova, and
  Xiuhong Zhang. 2014.
\newblock Not an interlingua, but close: Comparison of english amrs to chinese
  and czech.
\newblock In \emph{LREC}, volume~14, pages 1765--1772. Reykjavik, Iceland.

\bibitem[{Zhang et~al.(2019{\natexlab{a}})Zhang, Ma, Duh, and
  Van~Durme}]{zhang2019amr}
Sheng Zhang, Xutai Ma, Kevin Duh, and Benjamin Van~Durme. 2019{\natexlab{a}}.
\newblock Amr parsing as sequence-to-graph transduction.
\newblock In \emph{Proceedings of the 57th Annual Meeting of the Association
  for Computational Linguistics}, pages 80--94.

\bibitem[{Zhang et~al.(2019{\natexlab{b}})Zhang, Ma, Duh, and
  Van~Durme}]{zhang2019broad}
Sheng Zhang, Xutai Ma, Kevin Duh, and Benjamin Van~Durme. 2019{\natexlab{b}}.
\newblock Broad-coverage semantic parsing as transduction.
\newblock \emph{arXiv preprint arXiv:1909.02607}.

\end{thebibliography}


\end{document}